\pdfoutput=1

\documentclass[11pt]{article}

\usepackage[]{emnlp2021}

\usepackage{times}
\usepackage{latexsym}

\usepackage[T1]{fontenc}

\usepackage[utf8]{inputenc}

\usepackage{microtype}

\usepackage{times}
\usepackage{epsfig}
\usepackage{graphicx}
\usepackage{amsmath}
\usepackage{amssymb}
\usepackage{multirow}
\usepackage{colortbl}
\usepackage{hhline}
\usepackage{ulem}
\usepackage{comment}
\usepackage{url}
\usepackage{makecell}
\usepackage{xcolor}
\usepackage{caption}
\usepackage{subcaption}
\usepackage{xspace}
\usepackage{booktabs}
\usepackage{cleveref}
\usepackage{rotating}
\definecolor{cadmiumgreen}{rgb}{0.0, 0.42, 0.24}

\newcommand{\langaligned}{language-aligned\xspace}
\newcommand{\goalaligned}{goal-oriented\xspace}

\newcommand{\Langaligned}{Language-aligned\xspace}

\newcommand{\LAW}{LAW\xspace}
\newcommand{\wayaccuracy}{Waypoint Accuracy\xspace}
\newcommand{\modelcma}{CMA\xspace}

\newcommand{\pano}{\texttt{pano}\xspace}
\newcommand{\step}{\texttt{step}\xspace}
\newcommand{\goal}{\texttt{goal}\xspace}
\newcommand{\navgraph}{nav-graph\xspace}
\newcommand{\shortest}{shortest\xspace}
\newcommand{\langoriented}{language-oriented\xspace}
\newcommand{\goalthenlangmix}{G$\xrightarrow{}$L\xspace}
\newcommand{\goalthengoalandlangmix}{G$\xrightarrow{}$G+L\xspace}
\newcommand{\goalthengoalorlangmix}{G$\xrightarrow{}$G-or-L\xspace}

\newcommand{\xhdr}[1]{\vspace{0pt}\noindent\textbf{#1}\xspace}

\definecolor{pastelgreen}{HTML}{ade1b6}
\definecolor{pastelred}{HTML}{eecec5}
\definecolor{pastelred2}{HTML}{fec2c3}
\definecolor{pastelblue}{HTML}{ccd4eb}
\definecolor{lime}{HTML}{ade1b6}
\definecolor{pink}{HTML}{eecec5}

\title{Language-Aligned Waypoint (LAW) Supervision for\\Vision-and-Language Navigation in Continuous Environments}

\author{Sonia Raychaudhuri$^{1}$ \qquad
  Saim Wani$^{2}$ \qquad Shivansh Patel$^{2}$  \\ {\bf Unnat Jain$^{3}$} \qquad {\bf Angel X. Chang$^{1}$}  \\
  $^{1}$Simon Fraser University \qquad $^{2}$IIT Kanpur \qquad $^{3}$UIUC \\
  \small{$^{1}$\texttt{\{sraychau,angelx\}@sfu.ca}} \\ \small{ $^{2}$\texttt{\{saimdwani,shivanshpatel35\}@gmail.com} \qquad $^{3}$\texttt{uj2@illinois.edu}} \\
  \url{https://3dlg-hcvc.github.io/LAW-VLNCE}
}

\begin{document}
\maketitle

\begin{abstract}
In the Vision-and-Language Navigation (VLN) task an embodied agent navigates a 3D environment, following natural language instructions.
A challenge in this task is how to handle `off the path' scenarios where an agent veers from a reference path.
Prior work supervises the agent with actions based on the shortest path from the agent’s location to the goal, but such \goalaligned supervision is often not in alignment with the instruction.
Furthermore, the evaluation metrics employed by prior work do not measure how much of a language instruction the agent is able to follow.
In this work, we propose a simple and effective \langaligned supervision scheme, and a new metric that measures the number of sub-instructions the agent has completed during navigation.

\end{abstract}
\section{Introduction}
\label{sec:intro}

Training agents to navigate in realistic environments based on natural language instructions is a step towards building robots that understand humans and can assist them in their daily chores.
\citet{anderson2018vision} introduced the Vision-and-Language Navigation (VLN) task, where
an agent navigates a 3D environment to follow natural language instructions.
Much of the prior work on VLN assumes a discrete navigation graph (nav-graph), where the agent teleports between graph nodes, both in indoor~\cite{anderson2018vision} and outdoor~\cite{chen2019touchdown,mehta2020retouchdown} settings.
\citet{krantz2020beyond} reformulated the VLN task to a continuous environment (VLN-CE) by lifting the discrete paths to continuous trajectories, bringing the task closer to real-world scenarios.

\begin{figure}[t]
\centering
\includegraphics[width=\linewidth]{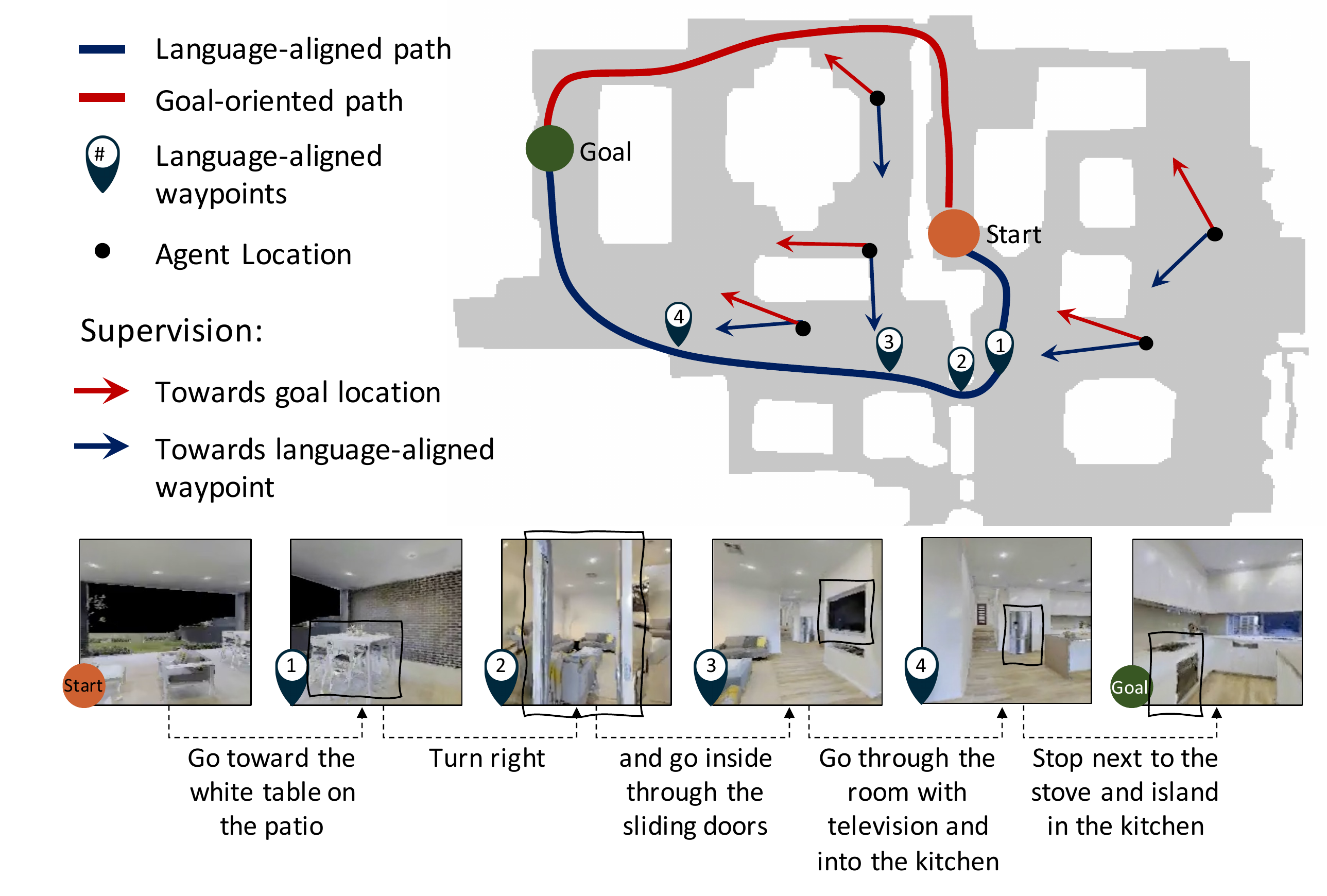}
\caption{
A \langaligned path (blue) in an instruction following task may differ from the shortest path (red) to the goal. \Langaligned supervision (blue arrows) encourages the agent at any given location (dark circles) to move towards the nearest waypoint on the \langaligned path and can hence be a better supervisory signal for instruction following than goal-oriented supervision (red arrows). 
}
\label{fig:teaser}
\end{figure}

\citet{krantz2020beyond} supervised agent training with actions based on the shortest path from the agent’s location to the goal, 
following prior work in VLN~\cite{fried2018speaker,tan2019learning,hu2019you,anderson2019chasing}.
However, as \citet{jain2019stay} observed, such supervision is \goalaligned and does not always correspond to following the natural language instruction.

Our key idea is that \langaligned supervision is better than \goalaligned supervision, as the path matching the instructions (\langaligned) may differ from the shortest path to the goal (\goalaligned).
This is especially true in `off the path' scenarios (where the agent veers off the reference path prescribed by the instructions).

Language-aligned supervision encourages the agent to move towards the nearest waypoint on the \langaligned path at every step and hence supervises the agent to better follow instructions (see \Cref{fig:teaser}).

In the discrete nav-graph setting, \citet{jain2019stay} interleave behavioral cloning and policy gradient training, with a sparse `fidelity-oriented reward' based on how well each node is covered on the reference path.
In contrast, we tackle the VLN-CE setting and propose a simple and effective approach that provides a denser supervisory signal leading the agent to the reference path.

A dense supervisory signal is especially important for VLN-CE where the episodes have a longer average length of 55.88 steps \textit{vs.} 4-6 nodes in (discrete) VLN. To this end, we conduct experiments investigating the effect of density of waypoint supervision on task performance.

To assess task performance, we complement the commonly employed normalized Dynamic Time Warping (nDTW) metric~\cite{ilharco2019general} with a intuitive \textit{Waypoint Accuracy} metric. Finally, to provide qualitative insights about degree of following instructions, we combine language-aligned waypoints with information about sub-instructions. Our experiments show that our \langaligned supervision trains agents to more closely follow instructions compared to \goalaligned supervision.
\section{Related Work}
\label{sec:rel}

\xhdr{Vision-and-Language Navigation.}
Since the introduction of the VLN task by \citet{anderson2018vision}, there has been a line of work exploring improved models and datasets.
The original Room-to-Room (R2R) dataset by \citet{anderson2018vision} provided instructions on a discrete navigation graph (nav-graph), with nodes corresponding to positions of panoramic cameras.
Much work focuses on this discrete nav-graph setting, including cross-modal grounding between language instructions and visual observations~\cite{wang2019reinforced}, addition of auxiliary progress monitoring~\cite{ma2019self}, augmenting training data by re-generating language instructions from trajectories~\cite{fried2018speaker}, and environmental dropout~\cite{tan2019learning}.

However, these methods fail to achieve similar performance in the more challenging VLN-CE task, where the agent navigates in a continuous 3D simulation environment.
\citet{chen2021topological} propose a modular approach using topological environment maps for VLN-CE and achieve better results. 
 
In concurrent work, \citet{krantz2021waypoint} propose a modular approach to predict waypoints on a panoramic observation space and use a low-level control module to navigate. However, both these works focus on improving the ability of the agent to reach the goal.
In this work, we focus on the VLN-CE task and on accurately following the path specified by the instruction.

\xhdr{Instruction Following in VLN.}
Work in the discrete nav-graph VLN setting has also focused on improving the agent's adherence to given instructions. \citet{anderson2019chasing} adopt Bayesian state tracking to model what a hypothetical human demonstrator would do when given the instruction, whereas \citet{qi2020object} attends to specific objects and actions mentioned in the instruction.
\citet{zhu2020babywalk} train the agent to follow shorter instructions and later generalize to longer instructions through a curriculum-based reinforcement learning approach.
\citet{hong2020sub} divide language instructions into shorter sub-instructions and enforce a sequential traversal through those sub-instructions.
They additionally enrich the Room-to-Room (R2R) dataset~\cite{anderson2018vision} with the sub-instruction-to-sub-path mapping and introduce the Fine-Grained R2R (FG-R2R) dataset.

More closely related to our work is \citet{jain2019stay}, which introduced a new metric -- Coverage weighted by Length Score (CLS), measuring the coverage of the reference path by the agent, and used it as a sparse fidelity-oriented reward for training. However, our work differs from theirs in a number of ways. First, in LAW we \textit{explicitly} supervise the agent to navigate back to the reference path, by dynamically calculating the closest waypoint (on the reference path) for any agent state. In contrast to calculating waypoints, \citet{jain2019stay} optimize accumulated rewards, based on the CLS metric. 
Moreover, we provide dense supervision (at every time step) for the agent to follow the reference path by providing a cross-entropy loss at all steps of the episode, in contrast to the single reward at the end of the episode during stage two of their training. Finally, LAW is an online imitation learning approach, which is simpler to implement and easier to optimize compared to their policy gradient formulation, especially with sparse rewards. Similar to \citet{jain2019stay}, \citet{ilharco2019general} train one of their agents with a fidelity oriented reward based on nDTW.

\section{Approach}
\label{sec:app}

\begin{figure}[t]
\centering
\includegraphics[width=\linewidth]{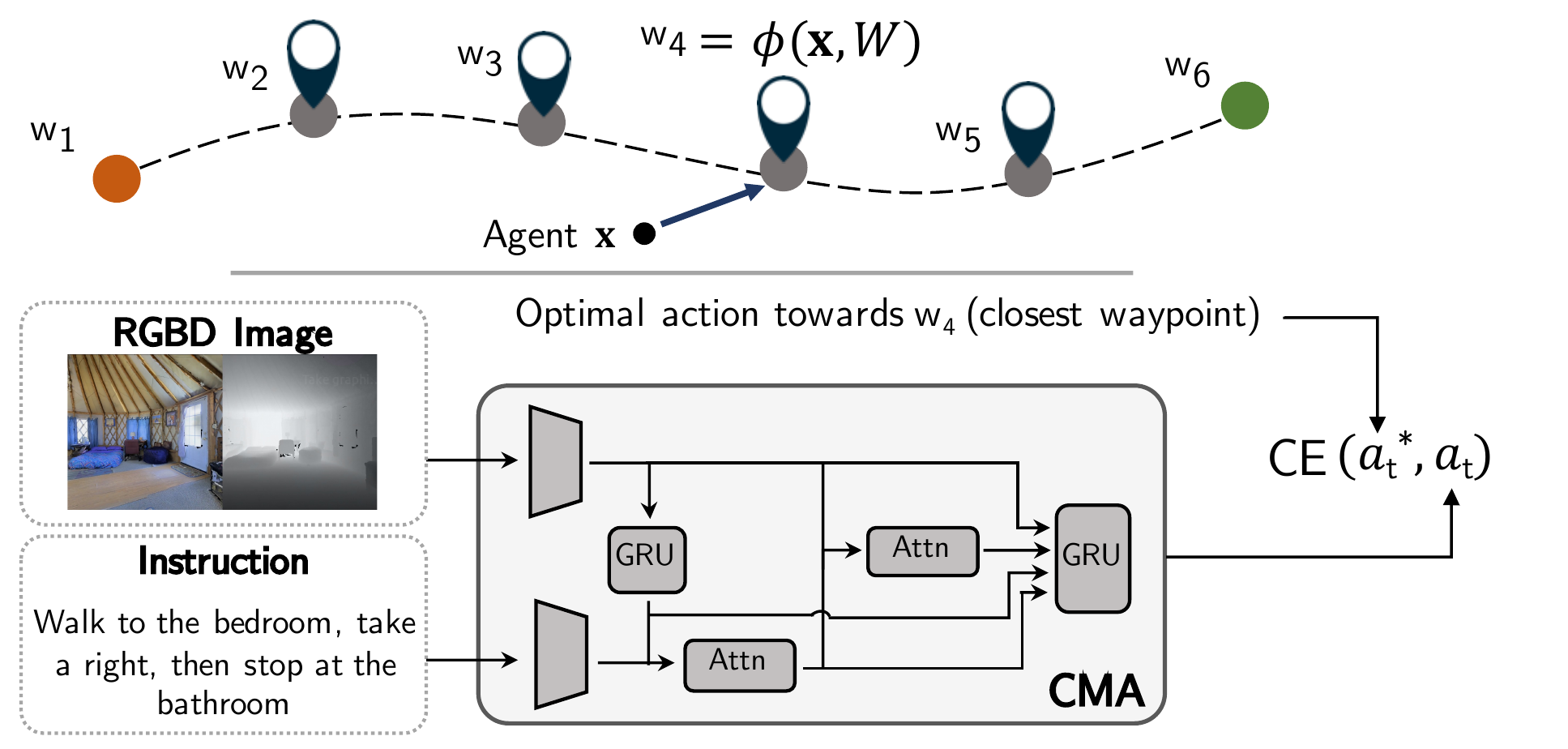}
\caption{ Top: The path from the start (orange) to the goal (green) with grey circle indicating \LAW\pano and the dashed segments indicating \LAW\step. Bottom: We adapt the  Cross-Modal Attention (CMA) model
~\citeyearpar{krantz2020beyond} which predicts an action. We optimize the model using \langaligned supervision, which brings it back on the path toward the next waypoint.}
\label{fig:model_diag}
\end{figure}

Our approach is evaluated on the VLN-CE dataset~\cite{krantz2020beyond}, which is generated by adapting R2R to the Habitat simulator~\cite{savva2019habitat}.
It consists of navigational episodes with language instructions and reference paths.
The reference paths are constructed by taking the discrete \navgraph nodes corresponding to positions of panoramic cameras (we call these \pano waypoints, shown as gray circles in \Cref{fig:model_diag} top), and taking the shortest geodesic distance between them to create a ground truth reference path consisting of dense waypoints  (\step waypoints, see dashed path in \Cref{fig:model_diag}) corresponding to an agent step size of $0.25m$.  

We take waypoints from these paths as \langaligned waypoints (\LAW) to supervise our agent, in contrast to the \goalaligned supervision in prior work.
We interpret our model performance qualitatively, and examine episodes for which the ground-truth \langaligned path (\LAW \step) does not match \goalaligned shortest path (\shortest)\footnote{We find ${\sim}6\%$ of the VLN-CE R2R episodes to have nDTW(\shortest, \LAW \step) $<0.8$ (see supplement)}.

\xhdr{Task.}
The agent is given a natural language instruction, and at each time step $t$, the agent observes the environment through RGBD image $I_t$ with a $90^{\circ}$ field-of-view, and takes one of four actions from $\mathcal{A}$: \{\textit{Forward}, \textit{Left}, \textit{Right}, \textit{Stop}\}. \textit{Left} and \textit{Right} turn the agent by $15^{\circ}$ and  \textit{Forward} moves forward by 0.25m. The \textit{Stop} action indicates that the agent has reached within a threshold distance of the goal.

\xhdr{Model.}
We adapt the Cross-Modal Attention (CMA) model (see \Cref{fig:model_diag}) which is shown to perform well on VLN-CE. It consists of two recurrent networks, one encoding a history of the agent state, and another predicting actions based on the attended visual and instruction features (see supplement for details).

\xhdr{Training.}
We follow the training regime of VLN-CE.
It involves two stages: behavior cloning (with teacher-forcing) on the larger augmented dataset to train an initial policy, and then fine-tuning with DAgger~\cite{ross2011reduction}. DAgger trains the model on an aggregated set of all past trajectories, sampling actions from the agent policy. Rather than supervising with the conventional \goalaligned sensor, we supervise with a \langaligned sensor in both the teacher-forcing phase and the DAgger phase.
The \langaligned sensor helps bring the agent back on the path to the next waypoint if it wanders off the path (see \Cref{fig:model_diag}~top).

The training dataset $\mathcal{D} = \{S^{(i)}, W^{(i)}\}$ consists of instructions $S^{(i)}$ and reference path $W^{(i)}$.
For each episode $(S,W) \sim \mathcal{D}$ with the agent starting state as $x_0$, we use cross entropy loss to maximize the log likelihood of the ground-truth action $a^*$ at each time step $t$:
$
\mathcal{L}_\text{CE}(x_0;\theta) = - \sum_{t=1}^T{\textbf{e}_{a^*}} \boldsymbol\cdot \log \pi_{\theta}(a_t | I_t, S, x_t, x_0)
$.
Here, 
$x_t$ is the 3D position of the agent at time $t$, $\textbf{e}_{a^*}$ is the one-hot vector for the ground-truth action $a^*$, which is defined as $a^* = g(x_t, \phi(x_t, W))$. The set of \langaligned waypoints is $W = \{w_1,...,w_m\}$. The waypoint in $W$ that is nearest to a 3D position $x_t$ is obtained by $\phi(x_t, W)$. The best action based on the shortest path from a 3D position $x_t$ to $w$ is denoted by $g(x_t, w)$.

\section{Experiments}
\label{sec:exp}

\xhdr{Dataset.}
We base our work on the VLN-CE dataset~\cite{krantz2020beyond}.
The dataset contains 4475 trajectories from Matterport3D~\cite{chang2017matterport3d}. 
Each trajectory is described by multiple natural language instructions.
The dataset also contains ${\sim}150k$ augmented trajectories generated by \citet{tan2019learning} adapted to VLN-CE.
To qualitatively analyze our model behavior, we use the Fine-Grained R2R (FG-R2R) dataset from \citealt{hong2020sub}.
It segments instructions into sub-instructions\footnote{There are on average 3.1 sub-instructions per instruction} and maps each sub-instruction to a corresponding sub-path.

\begin{figure}[t]
\centering
\includegraphics[width=0.95\linewidth]{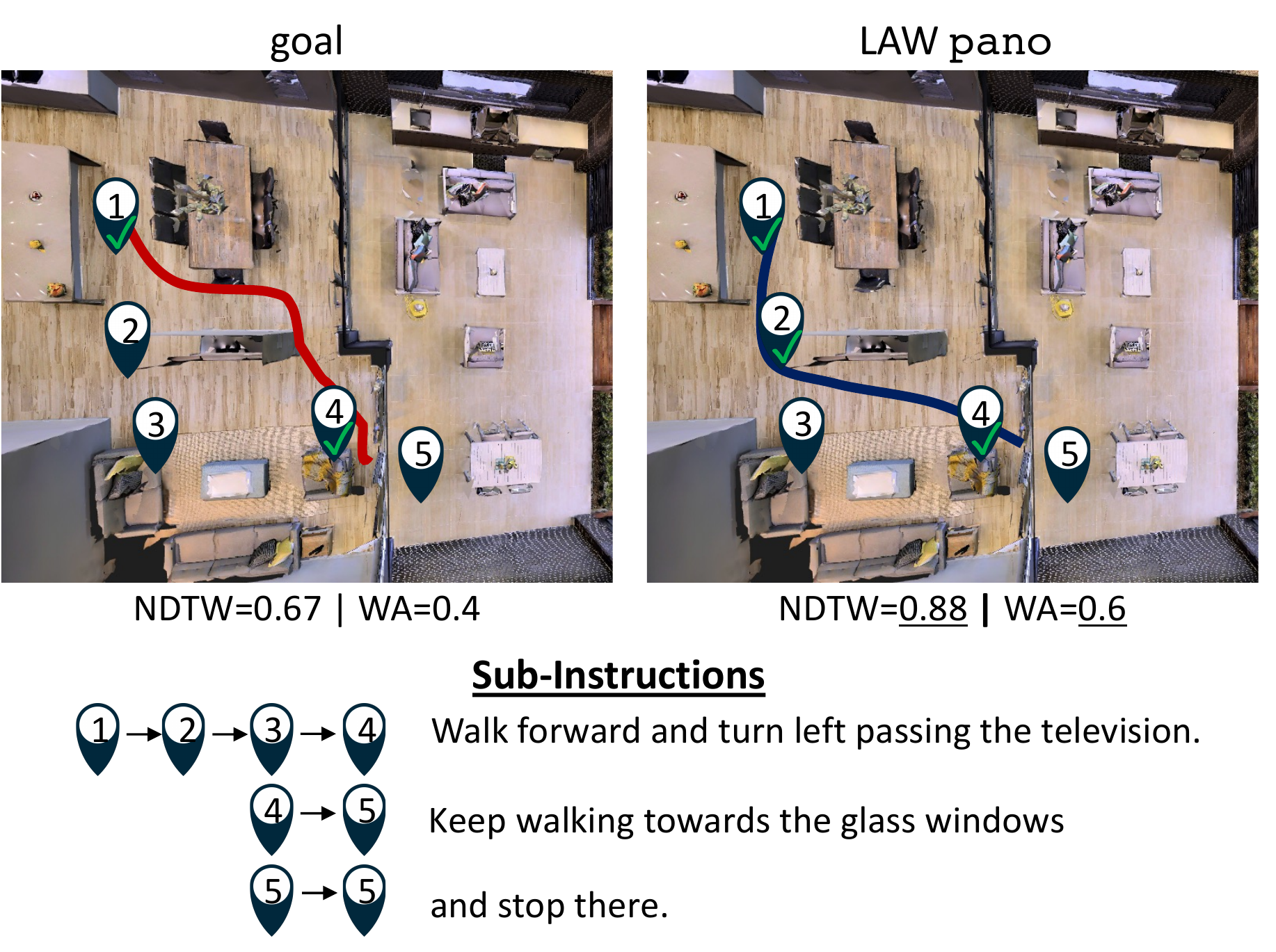}\vspace{-2mm}
\caption{
An example episode from R2R unseen split. The agent is able to learn to follow instruction better when supervised with \langaligned path (right) than the \goalaligned path (left). This is reflected in higher nDTW and waypoint accuracy (WA) metrics. Note that  WA can be intuitively visualized and interpreted.  We also show the mapping of sub-instructions to waypoints utilizing FG-R2R for this episode.
}
\label{fig:qual_anal_single_episode}
\end{figure}

\xhdr{Evaluation Metrics.}
We adopt standard metrics used by prior work~\cite{anderson2018vision,anderson2018evaluation,krantz2020beyond}.
In the main paper, we report Success Rate (SR), Success weighted by inverse Path Length (SPL), Normalized dynamic-time warping (nDTW), and Success weighted by nDTW (SDTW).
Trajectory Length (TL), Navigation Error (NE) and Oracle Success Rate (OS) are reported in the supplement.
Since none of the existing metrics directly measure how effectively waypoints are visited by the agent, we introduce \textit{\wayaccuracy} (WA) metric. It measures the fraction of waypoints the agent is able to visit correctly (specifically, within $0.5$m of the waypoint). This allows the community to intuitively analyze the agent trajectory as we illustrate in \Cref{fig:qual_anal_single_episode}.

\xhdr{Implementation Details.}
We implement our agents using PyTorch~\cite{paszke2019pytorch} and the Habitat simulator~\cite{savva2019habitat}. We build our code on top of the VLN-CE codebase\footnote{\url{https://github.com/jacobkrantz/VLN-CE}} and use the same set of hyper-parameters as used in the VLN-CE paper. The first phase of training with teacher forcing on the 150k augmented trajectories took ${\sim}$60 hours to train, while the second phase of training with DAgger on the original 4475 trajectories took ${\sim}$36 hours over two NVIDIA V100 GPUs.

\xhdr{Ablations.}
We study the effect of varying the density of \langaligned waypoints on model performance.
For all the ablations we use the CMA model described in \Cref{sec:app}.
We use \LAW\# to distinguish among the ablations.
On one end of the density spectrum, we have the base model which is supervised with only the goal (LAW\#1 or \goal).
On the other end is \LAW\step which refers to the pre-computed dense path from the VLN-CE dataset and can be thought of as the densest supervision available to the agent.
In the middle of the spectrum, we have \LAW\pano, which uses the navigational nodes (an average of 6 nodes) from the R2R dataset.
We also sample equidistant points on the \langaligned path to come up with LAW\#2, LAW\#4 and LAW\#15 containing two, four and fifteen waypoints, respectively.
  
The intuition is that \LAW\pano takes the agent back to the \langaligned path some distance ahead of its position, while \LAW\step brings it directly to the path.

\begin{table*}
\centering
\resizebox{0.8\linewidth}{!}{
\begin{tabular}{@{}lcccccccccccc@{}} 
\toprule
\multirow{2}{*}{Training} & & \multicolumn{5}{c}{\textbf{Val-Seen}} & & \multicolumn{5}{c}{\textbf{Val-Unseen}}\\ 
\cmidrule(lr){3-7}\cmidrule(lr){9-13}
& & SR$\uparrow$ & SPL$\uparrow$ & nDTW$\uparrow$ & sDTW$\uparrow$ & WA$\uparrow$ & & SR$\uparrow$ & SPL$\uparrow$  & nDTW$\uparrow$ & sDTW$\uparrow$ & WA$\uparrow$ \\
\midrule

\goal & &0.34 &0.32  & 0.54 & 0.29 &0.48 &
& 0.29	& 0.27	 & 0.50 &0.24 &0.41 \\

\LAW\pano & & \textbf{0.40} & \textbf{0.37} &  \textbf{0.58}  & \textbf{0.35} & \textbf{0.56} &
& \textbf{0.35}  & \textbf{0.31} &  \textbf{0.54} & \textbf{0.29} & \textbf{0.47} \\

\bottomrule
\end{tabular}}
\caption{\label{tab-short-vs-ref-path_short}
\textbf{Goal only vs \langaligned waypoint (LAW) supervision.} \LAW\pano performs better than \goal across all metrics, including the instruction-following metrics, nDTW and \wayaccuracy. This suggests that \langaligned supervision encourages the agent to follow instructions better than \goalaligned supervision.
}
\end{table*}

\begin{table*}
\centering
\resizebox{0.8\linewidth}{!}{
\begin{tabular}{@{}lcccccccccccc@{}} 
\toprule

\multirow{2}{*}{\LAW} & \multirow{2}{*}{\thead{Distance\\between\\waypoints}} & & \multicolumn{4}{c} {\textbf{Val-Seen}}                      & & \multicolumn{4}{c}{\textbf{Val-Unseen}}                      \\ 
\cmidrule(lr){4-7}\cmidrule(lr){9-12}
 & & & SR$\uparrow$ & SPL$\uparrow$ & nDTW$\uparrow$ & sDTW$\uparrow$ & & SR$\uparrow$ & SPL$\uparrow$ & nDTW$\uparrow$ & sDTW$\uparrow$ \\
\midrule

\#2 &5.00m & &0.39 &0.36 &0.57 &0.34 & &0.33 &0.30 &0.52 &0.28    \\

\#4 &2.50m & &0.35 &0.33 &0.54 &0.30  & &0.34 &0.31 &0.53 &0.29   \\

\pano(6) & 2.00m & & \textbf{0.40} & \textbf{0.37}  & \textbf{0.58} & \textbf{0.35}  &
& \textbf{0.35}  & \textbf{0.31}   & \textbf{0.54} & \textbf{0.29}  \\

\#15 &0.60m & &0.34 &0.32 &0.54 &0.29 & &0.33 &0.30 &0.52 &0.28    \\

\step & 0.25m & &0.37 &0.35  & 0.57 &0.32 & &0.32 &0.30  & 0.53  &0.27 \\
\bottomrule
\end{tabular}}
\caption{\label{tab-sparse-vs-dense-ref-path_short}
\textbf{Varying density of \langaligned supervision from very sparse (\#2) to dense (\step).}
This study shows that with varying density of the \langaligned waypoint supervision, the agent performs similarly, since all of them essentially follow the same path.
}
\end{table*}

\begin{figure}[t]
\centering
    \includegraphics[width=\linewidth]{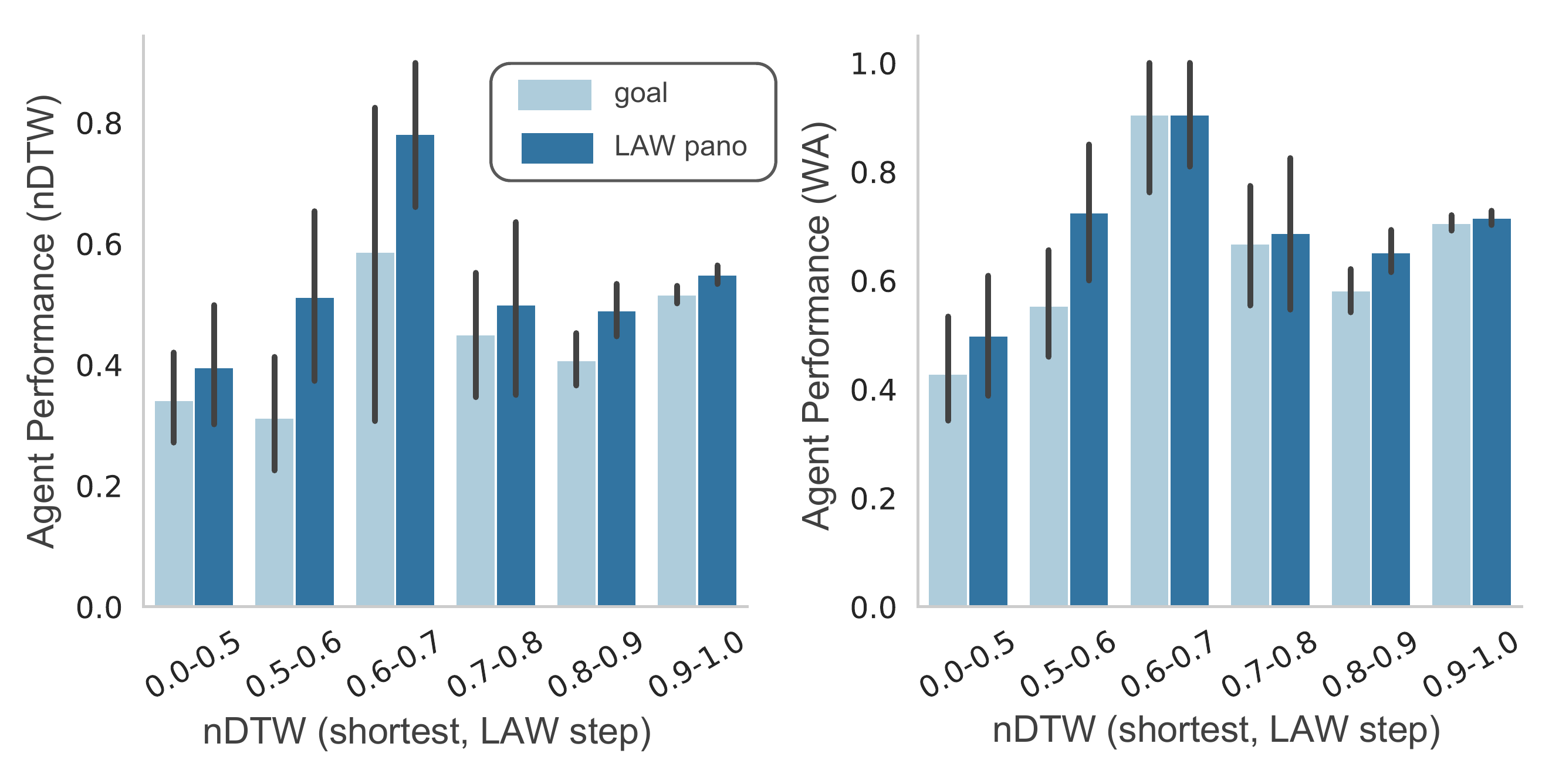}
\caption{
Agent performance binned by nDTW value of reference path to shortest path ($95\%$ CI error bars) shows that \LAW\pano performs better than \goal, especially on lower-range NDTW episodes. This indicates that \langaligned supervision is better suited for the instruction following task.
}
\label{fig:plot_langway_does_better}
\end{figure}

\xhdr{Quantitative Results.}
In \Cref{tab-short-vs-ref-path_short}, we see that \LAW\pano, supervised with the \langaligned path performs better than the base model supervised with the \goalaligned path, across all metrics in both validation seen and unseen environments.
We observe the same trend in the Waypoint Accuracy (WA) metric that we introduced.
\Cref{tab-sparse-vs-dense-ref-path_short} shows that agents perform similarly even when we vary the number of waypoints for the \langaligned path supervision, since all of them essentially follow the path.
This could be due to relatively short trajectory length in the R2R dataset (average of 10m) making \LAW \pano denser than needed for the instructions.
To check this, we analyze the sub-instruction data and find that one sub-instruction (e.g. `Climb up the stairs') often maps to several \pano waypoints, suggesting fewer waypoints are sufficient to specify the \langaligned path. For such paths, we find that the \LAW\#4 is better than the \LAW\pano (see supplement for details).

\Cref{fig:plot_langway_does_better} further analyzes model performance by grouping episodes based on similarity between the \goalaligned shortest path and the \langaligned path in the ground truth trajectories (measured by nDTW).
We find that the \LAW\space model performs better than the \goalaligned model, especially on episodes with dissimilar paths (lower nDTW) across both the nDTW and Waypoint Accuracy metrics.

\xhdr{Qualitative Analysis.}
To interpret model performance concretely with respect to path alignment we use the FG-R2R data, which contains mapping between sub-instructions and waypoints.
\Cref{fig:qual_anal_single_episode} contrast the agent trajectories of the \LAW\pano and \goalaligned agents on an unseen scene.
We observe that the path taken by the \LAW agent conforms more closely to the instructions (also indicated by higher nDTW). We present more examples in the supplement.

\xhdr{Additional Experiments.}
We additionally experiment with mixing \goalaligned and \langoriented losses while training, but observe that they fail to outperform the \LAW\pano model. The best performing mixture model achieves 53\% nDTW in unseen environment, as compared to 54\% nDTW for \LAW\pano (see supplement). Moreover, we perform a set of experiments on the recently introduced VLN-CE RxR dataset and observe that \langaligned supervision is better than \goalaligned supervision for this dataset as well, with \LAW\step showing a 6\%  increase  in  WA  and 2\%  increase  in  nDTW over \goal on the unseen environment. We defer the implementation details and results to the supplement.

\section{Conclusion}
\label{sec:conc}

We show that instruction following during the VLN task can be improved using \langaligned supervision instead of \goalaligned supervision as commonly employed in prior work.

Our quantitative and qualitative results demonstrate the benefit of the \LAW supervision.
The waypoint accuracy metric we introduce also makes it easier to interpret how agent navigation corresponds to following sub-instructions in the input natural language.
We believe that our results show that \LAW is a simple but useful strategy to improving VLN-CE. 

\paragraph{Acknowledgements}
We thank Jacob Krantz for the VLN-CE code on which this project was based, Erik Wijmans for initial guidance with reproducing the original VLN-CE results, and Manolis Savva for discussions and feedback.  We also thank the anonymous reviewers for their suggestions and feedback.  This work was funded in part by a Canada CIFAR AI Chair and NSERC Discovery Grant, and enabled in part by support provided by \href{www.westgrid.ca}{WestGrid} and \href{www.computecanada.ca}{Compute Canada}. 

\bibliographystyle{acl_natbib}
\bibliography{main}
\clearpage
\appendix

\begin{appendix}
\section{Appendix}

\subsection{Glossary}
Some commonly used terminologies in this work are described here:

\begin{itemize}

    \item \LAW  refers to \langaligned waypoints such that the navigation path aligns with the language instruction.
    \item \navgraph  refers to the discrete navigation graph of a scene.
    \item \pano refers to the reference paths constructed by taking the discrete \navgraph nodes corresponding to positions of panoramic cameras in the R2R dataset.
    \item \step refers to the reference paths constructed by taking the shortest geodesic distance between the \pano paths to create dense waypoints corresponding to an agent step size of $0.25m$.
    \item `\shortest'  refers to the \goalaligned path, i.e, the shortest path to the goal.
    \item \goal refers to the model supervised with only the goal.

\end{itemize}

\subsection{Analysis of VLN-CE R2R path}

We analyze the similarity of the VLN-CE R2R reference path to the shortest path using nDTW.  We find that ${\sim}6\%$ of episodes (including training and validation splits), have nDTW(shortest, \LAW\step)~$< 0.8$.

\begin{figure*}[t]
\centering
\includegraphics[width=\linewidth]{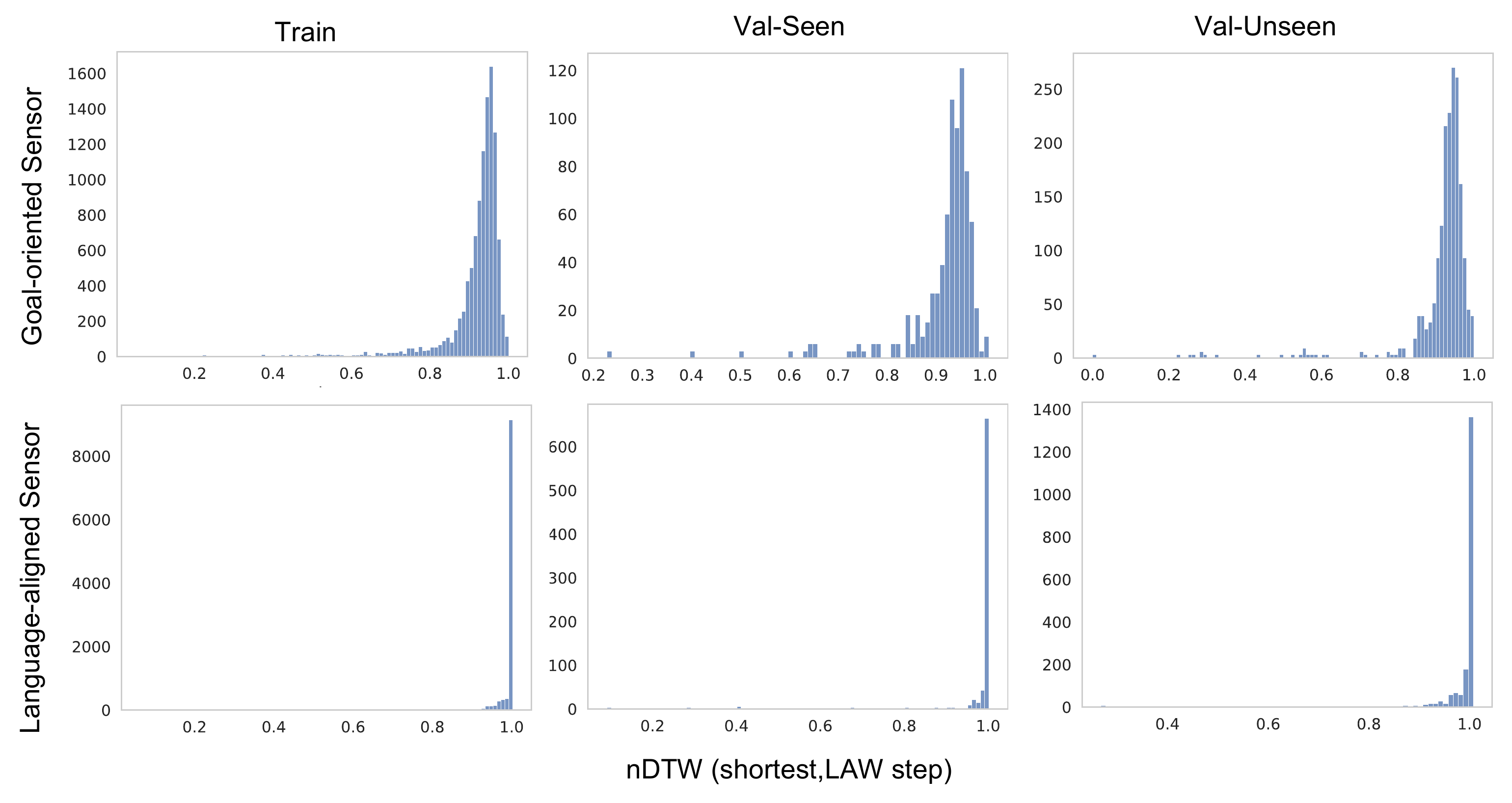}
\caption{
Plots showing a distribution of the number of R2R episodes across different nDTW values of reference path to shortest path for train, val-seen and val-unseen splits. There are many episodes for which the \goalaligned shortest path does not match the \langaligned path, as generated by the \goalaligned action sensor (top). We mitigate this problem by using \langaligned action sensor (bottom).
}
\label{fig:plot_langway_better_ndtw}
\end{figure*}

\begin{figure}[t]
\centering
\includegraphics[width=\linewidth]{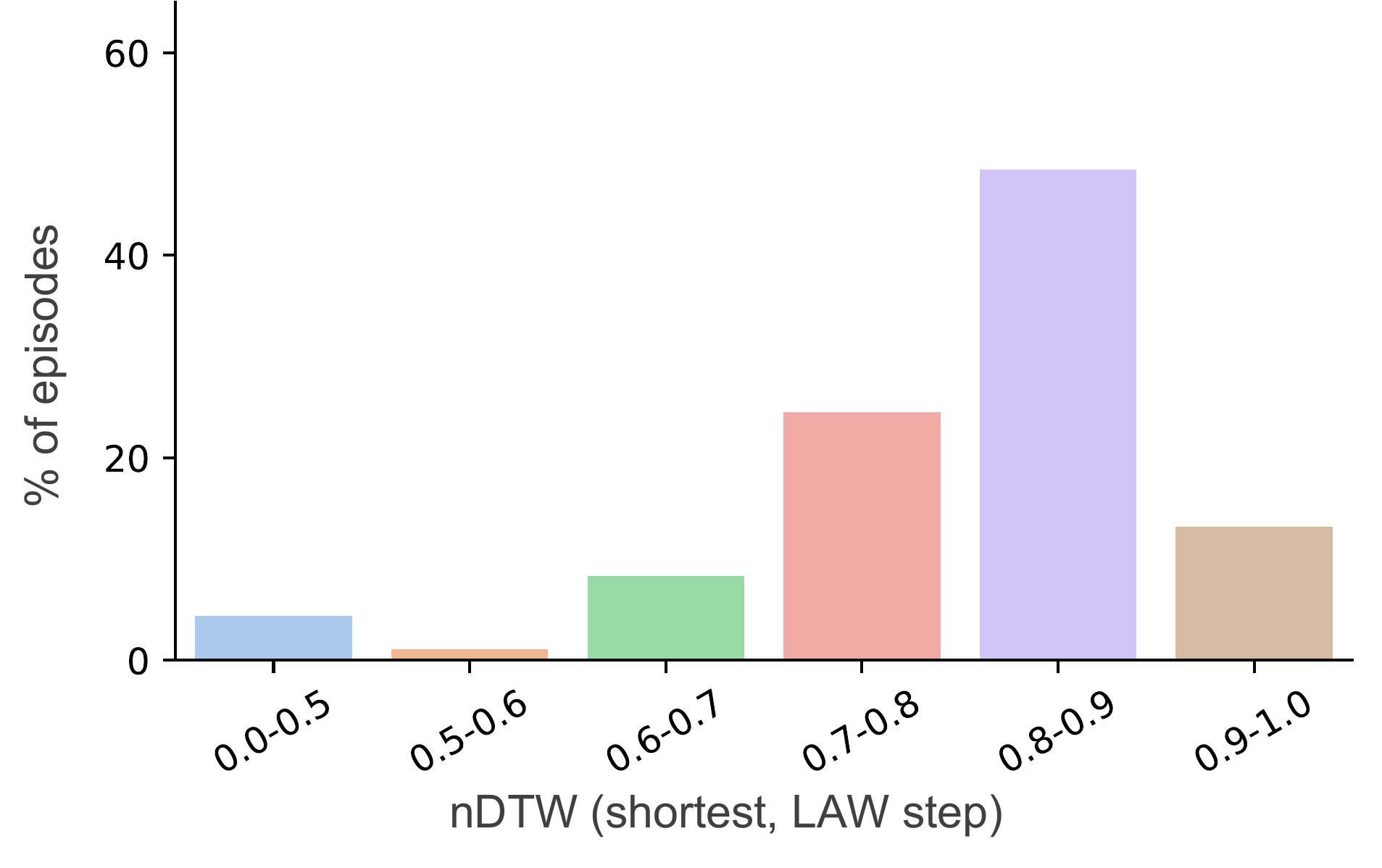}
\caption{
Plot showing percentage of episodes in R2R unseen dataset binned by nDTW value of reference path to shortest path. The lower nDTW values indicate episodes for which the \goalaligned shortest path does not match \langaligned path.
}
\label{fig:suppl_percent_episodes}
\end{figure}

\begin{figure}[t]
\centering
\includegraphics[width=\linewidth]{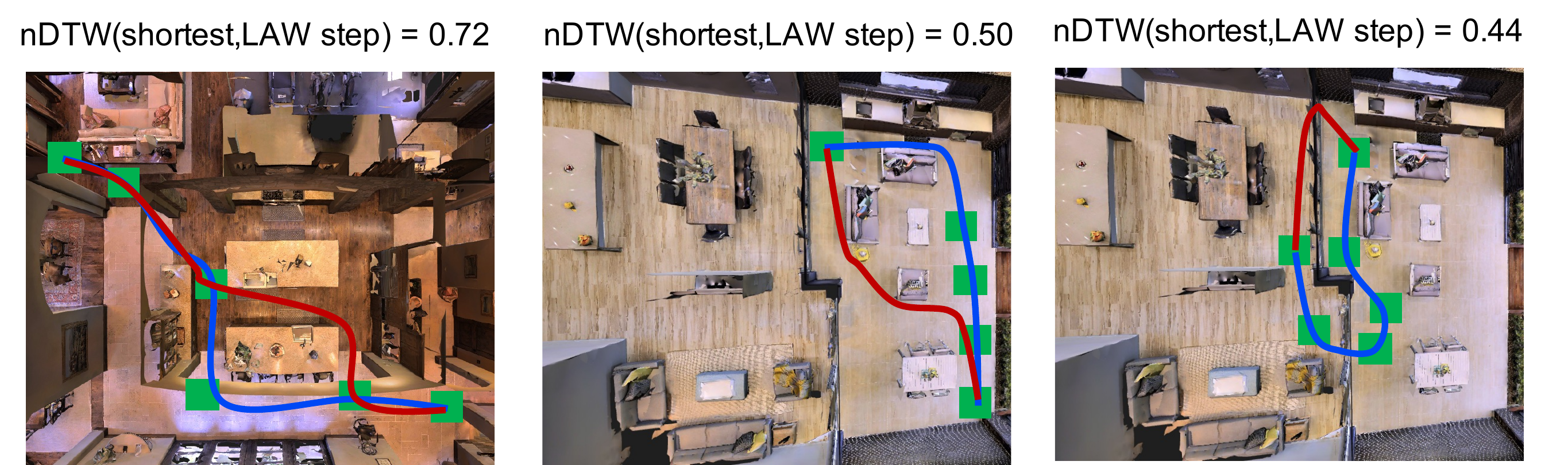}
\caption{
Visualization of a few R2R unseen episodes with nDTW(\shortest,\LAW \step)$<0.8$ shows us how dissimilar the \goalaligned shortest path (red) and the \langaligned path (blue) are.
}
\label{fig:suppl_ndtw_example}
\end{figure}

\Cref{fig:plot_langway_better_ndtw} shows the distribution of nDTW of the ground truth trajectories (LAW \step) against the shortest path (\goalaligned action sensor) and LAW \pano (\langaligned action sensor). It shows that the two distributions are different and that the \langaligned sensor will be much closer to the ground truth trajectories. \Cref{fig:suppl_percent_episodes} shows percentage of unseen episodes binned  by  nDTW  value  of  reference path to shortest path, which helps us analyze our model performance as shown in \Cref{fig:plot_langway_does_better} (main paper). Additionally, we visualize a few such paths to see how dissimilar they are in \Cref{fig:suppl_ndtw_example}.

\subsection{CMA Model}
The Cross-Modal Attention (CMA) Model takes the input RGB and depth observations and encode them using a ResNet50~\cite{he2016deep} pre-trained on ImageNet~\cite{deng2009imagenet} and a modified ResNet50 trained on point-goal navigation~\cite{wijmans2019dd} respectively.
It also takes as input the GLoVE~\cite{pennington2014glove} embeddings for the tokenized words in the language instruction and pass them through a bi-directional LSTM to obtain their feature representations. 
The CMA model consists of two recurrent (GRU) networks. The first GRU encodes a history of the agent state, which is then used to generate attended instruction features. These attended instruction features are in turn used to generate visual attentions. The second GRU takes in all the features generated thus far to predict an action. The attention used here is a scaled dot-product attention~\cite{vaswani2017attention}.

\subsection{Full Evaluation}

\begin{table*}
\centering
\resizebox{\linewidth}{!}{
\begin{tabular}{lccccccccccccccccc} 
\toprule
\multirow{2}{*}{\LAW} &\multirow{2}{*}{\thead{Distance\\between\\waypoints}} & \multicolumn{8}{c} {\textbf{Val-Seen}}                       & \multicolumn{8}{c}{\textbf{Val-Unseen}}                      \\ 
\cmidrule(lr){3-10}\cmidrule(lr){11-18}
& & TL$\downarrow$ & NE$\downarrow$ & OS$\uparrow$ & SR$\uparrow$ & SPL$\uparrow$ & nDTW$\uparrow$ & sDTW$\uparrow$ & WA$\uparrow$  & TL$\downarrow$ & NE$\downarrow$ & OS$\uparrow$ & SR$\uparrow$ & SPL$\uparrow$ & nDTW$\uparrow$ & sDTW$\uparrow$ & WA$\uparrow$  \\
\midrule

\#1(\goal) &10m & 9.06 & 7.21 & 0.44 & 0.34 & 0.32 & 0.54 &0.29 &0.48
& 8.27	& 7.60	& 0.36	& 0.29	& 0.27 & 0.50 &0.24 & 0.41 \\

\midrule

\#2 &5m &9.39 &6.76 &0.47 &0.39 &0.36 &0.57 &0.34 &0.51 &8.57 &7.41 &0.39 &0.33 &0.30 &0.52 &0.28  &0.44  \\

\#4 &2.5m &9.08 &6.94 &0.47 &0.35 &0.33 &0.54 &0.30 &0.49 &8.57 &7.01 &0.41 &0.34 &0.31 &0.53 &0.29  &0.44 \\

\pano & 2m & 9.34 & 6.35 & 0.49 & \textbf{0.40} & \textbf{0.37} & \textbf{0.58} & \textbf{0.35}  & \textbf{0.56}  
& 8.89  & 6.83 & 0.44 & \textbf{0.35} & \textbf{0.31} & \textbf{0.54} & \textbf{0.29} & \textbf{0.47}   \\

\#15 &0.6m &9.51 &7.16 &0.45 &0.34 &0.32 &0.54 &0.29 &0.50 &8.71 &7.05 &0.41 &0.33 &0.30 &0.52 &0.28  &0.44      \\

\step & 0.25m &9.76 &6.35  & 0.49 & 0.37 & 0.35 & 0.57 &0.32 &0.50 &9.06 &6.81 & 0.40 & 0.32 & 0.30 & 0.53  &0.27 &0.44 \\

\bottomrule
\end{tabular}}
\caption{\label{tab-short-vs-ref-path_full}
\LAW\pano model supervised with \langaligned waypoints performs better than the same model supervised with \goalaligned path, i.e. the shortest path to the goal. All models supervised with \langaligned path, but with varying density, perform similarly.
}
\end{table*}

\subsubsection{Metrics}
We report the full evaluation of the models here on the standard metrics for VLN such as:

\xhdr{Trajectory Length (TL)}: agent trajectory length.

\xhdr{Navigation Error (NE)}: distance from agent to goal at episode termination.

\xhdr{Success Rate (SR)}: rate of agent stopping within a threshold distance (around 3 meters) of the goal.

\xhdr{Oracle Success Rate (OS)}: rate of agent reaching within a threshold distance (around 3 meters) of the goal at any point during navigation.

\xhdr{Success weighted by inverse Path Length (SPL)}: success weighted by trajectory length relative to shortest path trajectory between start and goal.

\xhdr{Normalized dynamic-time warping (nDTW)}: evaluates how well the agent trajectory matches the ground truth trajectory.

\xhdr{Success weighted by nDTW (SDTW)}: nDTW, but calculated only for successful episodes.

\subsubsection{Quantitative Results}
\begin{figure}[t]
\centering
\includegraphics[width=\linewidth]{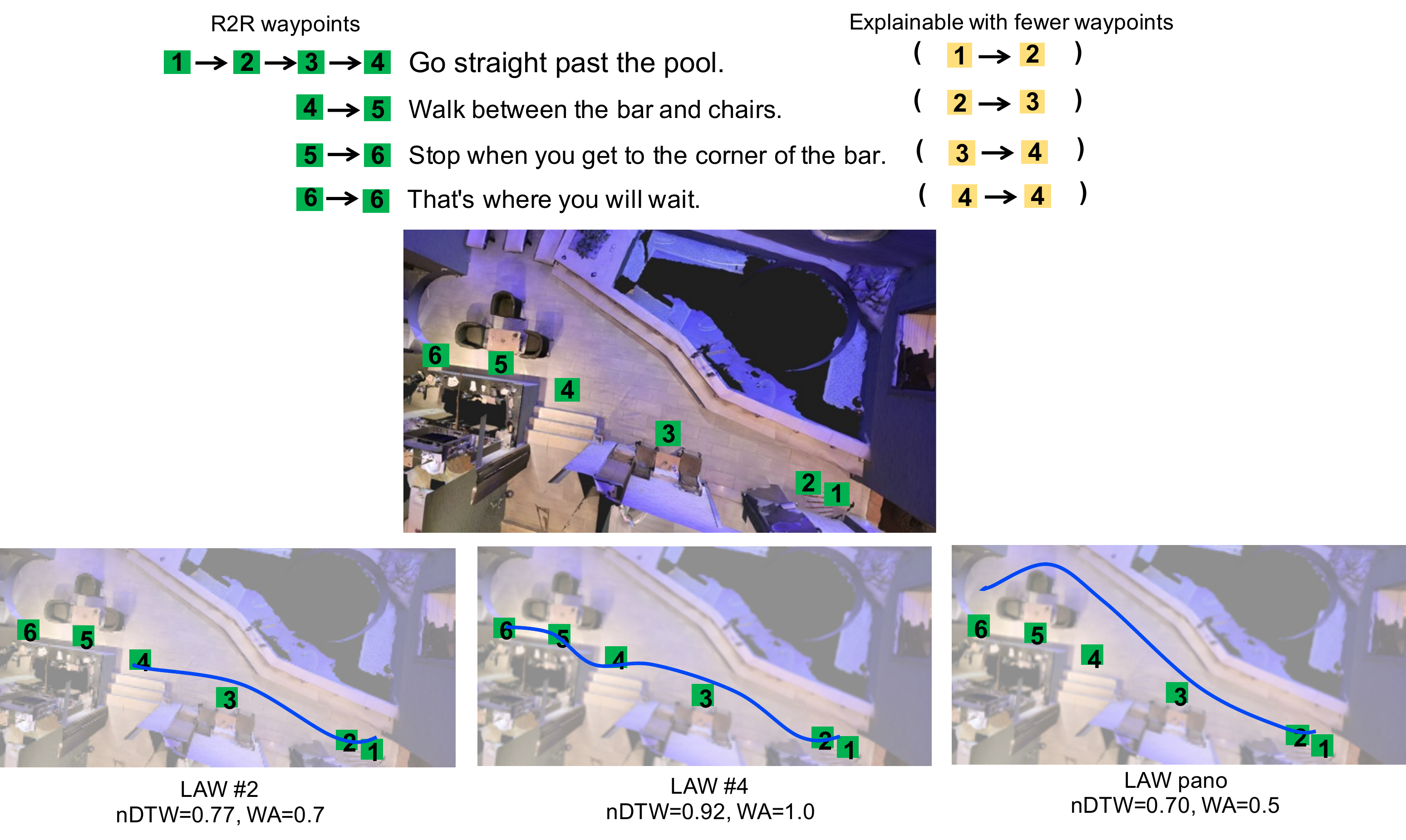} 
\caption{Top: Analysis of the R2R dataset along with the FG-R2R shows that one sub-instruction often maps to several \pano waypoints and can be explained via fewer waypoints.\\
Bottom: The above path can be defined via 4 waypoints and evaluating the episode with the different model variations shows that \LAW\#4 (supervising with 4 waypoints) performs the best.}
\label{fig:suppl_density_res}
\end{figure}

We observe that the models in the ablation (\LAW\#2 to \LAW\step in Table~\ref{tab-short-vs-ref-path_full}) perform similarly, which could be due to the fact that the average trajectory length in the R2R dataset is around 10m and the \LAW \pano is actually denser than the agent needs to follow instructions. We analyze this by using the sub-instruction data and find that one sub-instruction often maps to several \pano waypoints and the \langaligned path can be explained via fewer waypoints. We show some such examples from the dataset in \Cref{fig:suppl_density_res}.
We also report the results on the R2R test split in \Cref{tab-short-vs-ref-on-test}, which shows that \LAW\pano performs better on OS, while performing similarly to \goal on SR and SPL metrics. However, since the VLN-CE leaderboard\footnote{https://eval.ai/web/challenges/challenge-page/719} does not report the instruction-following metrics, nDTW and sDTW, we could not report how well the \LAW\pano agent follows instructions on the test set.

\begin{table}
\centering
\resizebox{0.8\linewidth}{!}{
\begin{tabular}{@{}lccccccc@{}} 
\toprule
\multirow{2}{*}{Training} & &  \multicolumn{5}{c}{\textbf{Test}}\\ 
\cmidrule(lr){3-7}
& & TL$\downarrow$ & NE$\downarrow$ & OS$\uparrow$ & SR$\uparrow$ & SPL$\uparrow$ \\
\midrule

\goal & & 8.85 & 7.91 & 0.36 & 0.28 & 0.25 \\

\LAW\pano  & & 9.67 & 7.69 & \textbf{0.38} & 0.28 & 0.25 \\

\bottomrule
\end{tabular}}
\caption{\label{tab-short-vs-ref-on-test}
Evaluating \LAW\pano on the VLN-CE test split gives us an increase in OS, although the SR and SPL is same as the \goal.
}
\end{table}

\subsubsection{Qualitative Analysis}
\Cref{tab-qual-analysis} shows a qualitative interpretation of some R2R unseen episodes for the two models \goal and \LAW\pano, along with the sub-instruction data from the FG-R2R dataset.  We see that \LAW\pano is able to get more number of waypoints (and hence sub-instructions) correct than the \goal model. We report the Waypoint Accuracy metric at threshold distances of 0.5m and 1.0m for the same. It also shows that Waypoint Accuracy is more intuitive than nDTW in terms of interpreting what fraction of waypoints the agent is able to predict correctly.

\begin{table*}
\resizebox{\linewidth}{!}{
\begin{tabular}{@{}ll@{}} 
\toprule
\textbf{\modelcma+~\goal} &  \textbf{\modelcma+~\LAW\pano} \\ 
\midrule

\makecell[l]{nDTW=0.52, WA@0.5m=0.4, WA@1.0m=0.8 \\\colorbox{lime}{1$\rightarrow$4 Walk straight through the living room towards the stairs.}
\\\colorbox{pink}{4$\rightarrow$5 Go to the right of the stairs towards the dining area }
\\\colorbox{pink}{5$\rightarrow$5 and wait by the leather chair at the entry to the dining room.}
}
& 
\makecell[l]{nDTW=0.98, WA@0.5m=1.0, WA@1.0m=1.0
\\\colorbox{lime}{1$\rightarrow$4 Walk straight through the living room towards the stairs.}
\\\colorbox{lime}{4$\rightarrow$5 Go to the right of the stairs towards the dining area }
\\\colorbox{lime}{5$\rightarrow$5 and wait by the leather chair at the entry to the dining room.}
}\\
\midrule

\makecell[l]{nDTW=0.15, WA@0.5m=0.4, WA@1.0m=0.6 \\\colorbox{lime}{1$\rightarrow$3 Walk straight into the kitchen area. }
\\\colorbox{pink}{3$\rightarrow$4 Turn left and exit the kitchen }
\\\colorbox{pink}{4$\rightarrow$5 and stop there. }
}
& \makecell[l]{nDTW=0.74, WA@0.5m=0.8, WA@1.0m=1.0 \\\colorbox{lime}{1$\rightarrow$3 Walk straight into the kitchen area. }
\\\colorbox{lime}{3$\rightarrow$4 Turn left and exit the kitchen }
\\\colorbox{lime}{4$\rightarrow$5 and stop there. } 
} \\
\midrule

\makecell[l]{nDTW=0.26, WA@0.5m=0.2, WA@1.0m=0.2 \\\colorbox{lime}{1$\rightarrow$4 Go to exit of writing room. } (1 correct)
\\\colorbox{pink}{4$\rightarrow$5 Stop between pillars.}
}
& \makecell[l]{nDTW=0.95, WA@0.5m=1.0, WA@1.0m=1.0
\\\colorbox{lime}{1$\rightarrow$4 Go to exit of writing room. }
\\\colorbox{lime}{4$\rightarrow$5 Stop between pillars.} 
} \\
\bottomrule
\end{tabular}
}
\caption{
We analyse model performance on sub-instruction data~\citeyearpar{hong2020sub}. \modelcma+~\LAW\pano (right) correctly predicts more sub-instructions compared to \modelcma+~\goal (left).
Mapping between sub-instruction and waypoints is indicated by start and end waypoint indices.
\colorbox{lime}{Green} and \colorbox{pink}{Red} indicate correct and incorrect prediction respectively. 
WA@0.5m and WA@1.0m indicate Waypoint Accuracy measured at a threshold distance of 0.5m and 1.0m respectively from the waypoint.
}
\label{tab-qual-analysis}
\end{table*}

\subsection{Mixing \goalaligned and \langoriented losses}

We experiment with mixing \goalaligned loss (G) and \langoriented loss (L) during training to further understand the contribution \langoriented supervision. We pre-trained with G using teacher forcing and then fine-tuned with (a) only L, (b) L+G, (c) randomly chosen L or G, using DAgger. The results as reported in Table~\ref{tab-goal-law-mix} show that none of the models outperform \LAW\pano, indicating that training with mixed losses fail to perform as well as training with only language-oriented loss.

\begin{table*}
\centering
\resizebox{\linewidth}{!}{
\begin{tabular}{lcccccccccccc} 
\toprule
\multirow{2}{*}{Training} & \multicolumn{6}{c} {\textbf{Val-Seen}}                       & \multicolumn{6}{c}{\textbf{Val-Unseen}}                      \\ 
\cmidrule(lr){2-7}\cmidrule(lr){8-13}
 & OS$\uparrow$ & SR$\uparrow$ & SPL$\uparrow$ & nDTW$\uparrow$ & sDTW$\uparrow$ & WA$\uparrow$  & OS$\uparrow$ & SR$\uparrow$ & SPL$\uparrow$ & nDTW$\uparrow$ & sDTW$\uparrow$ & WA$\uparrow$  \\
\midrule

\LAW\pano & \textbf{0.49} & \textbf{0.40} & \textbf{0.37} & \textbf{0.58} & \textbf{0.35}  & \textbf{0.56}  
& \textbf{0.44} & \textbf{0.35} & \textbf{0.31} & \textbf{0.54} & \textbf{0.29} & \textbf{0.47}   \\

\midrule

\goalthenlangmix & 0.49 & 0.36  & 0.34 & 0.56 & 0.32 & 0.56  
& 0.40 & 0.32  & 0.29 & 0.51 & 0.27 & 0.47  \\

\goalthengoalandlangmix & 0.46 & 0.34  & 0.31 & 0.57 & 0.30 & 0.55 
& 0.38 & 0.27 & 0.25 & 0.51 & 0.23 & 0.45  \\

\goalthengoalorlangmix & 0.45 & 0.35  & 0.34 & 0.55 & 0.31 & 0.51  
& 0.41 & 0.32 & 0.29 & 0.53 & 0.27 & 0.46 \\

\bottomrule
\end{tabular}}
\caption{\label{tab-goal-law-mix}
Experiments show that models trained with a mixture of \goalaligned (G) and \langoriented (L) superivision underperforms the model trained with only our \langoriented loss.
}
\end{table*}

\begin{table*}
\centering
\resizebox{\linewidth}{!}{
\begin{tabular}{lcccccccccccccccc} 
\toprule
\multirow{2}{*}{Training} & \multicolumn{8}{c} {\textbf{Val-Seen}}                       & \multicolumn{8}{c}{\textbf{Val-Unseen}}                      \\ 
\cmidrule(lr){2-9}\cmidrule(lr){10-17}
 & TL$\downarrow$ & NE$\downarrow$ & OS$\uparrow$ & SR$\uparrow$ & SPL$\uparrow$ & nDTW$\uparrow$ & sDTW$\uparrow$ & WA$\uparrow$  & TL$\downarrow$ & NE$\downarrow$ & OS$\uparrow$ & SR$\uparrow$ & SPL$\uparrow$ & nDTW$\uparrow$ & sDTW$\uparrow$ & WA$\uparrow$  \\
\midrule

\goal & 4.55 & 12.01  & 0.14 & 0.06 & 0.05 & 0.34  & 0.05 & 0.28
& 4.03 & 11.00  & 0.16 & 0.06 & 0.05 & 0.36  & 0.05 & 0.27  \\

\LAW\pano & 6.27 & 12.07  & 0.17 & \textbf{0.09} & \textbf{0.09} & 0.35  & 0.08    & 0.31
& 4.62 & 11.04  & 0.16 & \textbf{0.10} & \textbf{0.09} & 0.37 & 0.08  & 0.28   \\

\LAW\step & 7.92 & 11.94 & 0.20 & 0.07 & 0.06 & \textbf{0.36} & 0.06 & \textbf{0.35} & 4.01 & 10.87 & 0.21 & 0.08 & 0.08 & \textbf{0.38} & 0.07 & \textbf{0.33}    \\

\bottomrule
\end{tabular}}
\caption{\label{tab-rxr}
Experiments on the recently released RxR-Habitat benchmark (English language split) show that \LAW methods outperform the \goal, with \LAW\step having a 6\% increase in WA and 2\% increase in nDTW over \goal on unseen environment. This indicates that our idea of \langaligned supervision is useful beyond R2R.
}
\end{table*}

\subsection{Evaluation on VLN-CE RxR dataset}
\xhdr{Dataset.} Beyond the R2R dataset, there exist VLN datasets where the aim is to have the \langaligned path not be the shortest path. \citet{jain2019stay} proposed the Room-for-Room (R4R) dataset by combining paths from R2R.
More recently, \citet{ku2020room} introduced the Room-Across-Room (RxR) dataset that consists of new trajectories designed to not match the shortest path between start and goal. Importantly, they do not have a bias on the path length itself. The RxR dataset is 10x larger than R2R and consists of longer trajectories and instructions in three languages, English, Hindi, and Telugu. Both the R4R and RxR datasets are on the discrete nav-graph setting. However, the RxR dataset has been recently ported to the continuous state space of the VLN-CE for the RxR-Habitat challenge at CVPR 2021\footnote{\url{https://github.com/jacobkrantz/VLN-CE/tree/rxr-habitat-challenge}}. We experiment on the VLN-CE RxR to further investigate if \langaligned supervision is better than \goalaligned supervision on a dataset other than R2R.

\xhdr{Model.} We build our experiments on the model architecture provided for the VLN-CE RxR Habitat challenge. The only difference in the CMA architecture in VLN-CE RxR from the one used in VLN-CE is that they use pre-computed BERT features for the language instructions instead of GLoVE embeddings. We vary the nature of supervision as before. The \goal model receives \goalaligned supervision, whereas the \LAW\pano and \LAW\step are supervised with \langaligned \pano and \step waypoints, respectively. The baseline model in the VLN-CE RxR codebase also follows the \step supervision.

\xhdr{Training.} We train the methods with teacher forcing as was done in the baseline model in the VLN-CE RxR Habitat challenge. However, we used only the RxR English language split for both training and evaluating our models. Note that the training regime here is different from that of our main R2R experiments and does not have the DAgger fine-tuning phase.

\xhdr{Results.} Table~\ref{tab-rxr} shows the results of our experiments. We observe that the \LAW methods outperform the \goal, with \LAW\step showing a 6\% increase in WA and 2\% increase in nDTW on the unseen environment. This indicates that language-aligned supervision is useful beyond the R2R dataset. 

\end{appendix}

\end{document}